\documentclass[sn-mathphys-num]{sn-jnl}

\usepackage{graphicx}%
\usepackage{multirow}%
\usepackage{amsmath,amssymb,amsfonts}%
\usepackage{amsthm}%
\usepackage{mathrsfs}%
\usepackage[title]{appendix}%
\usepackage{xcolor}%
\usepackage{textcomp}%
\usepackage{manyfoot}%
\usepackage{booktabs}%
\usepackage{algorithm}%
\usepackage{algorithmicx}%
\usepackage{algpseudocode}%
\usepackage{listings}%
\usepackage{xcolor}
\usepackage{soul}

\theoremstyle{thmstyleone}%
\theoremstyle{thmstyletwo}%

\theoremstyle{thmstylethree}%

\begin{document}

\title[Article Title]{A Survey for Foundation Models in Autonomous Driving}


\author[1]{\fnm{Haoxiang} \sur{Gao}}

\author[2]{\fnm{Zhongruo} \sur{Wang}}

\author[3]{\fnm{Yaqian} \sur{Li}}

\author[3]{\fnm{Kaiwen} \sur{Long}}

\author[4]{\fnm{Ming} \sur{Yang}}

\author*[5]{\fnm{Yiqing} \sur{Shen}}\email{yshen92@jhu.edu}


\affil[1]{\orgname{Motional AD LLC}}

\affil[2]{\orgname{Amazon}}

\affil[3]{\orgdiv{Li Auto Inc.}}

\affil[4]{\orgname{Shanghai Jiaotong University}}

\affil[5]{\orgname{John Hopkins University}}


\abstract{The advent of foundation models has revolutionized the fields of natural language processing and computer vision, paving the way for their application in autonomous driving (AD).
    This survey presents a comprehensive review of more than 40 research papers, demonstrating the role of foundation models in enhancing AD.
    Large language models contribute to planning and simulation in AD, particularly through their proficiency in reasoning, code generation and translation. 
    In parallel, vision foundation models are increasingly adapted for critical tasks such as 3D object detection and tracking, as well as creating realistic driving scenarios for simulation and testing.
    Multi-modal foundation models, integrating diverse inputs, exhibit exceptional visual understanding and spatial reasoning, crucial for end-to-end AD. 
    This survey not only provides a structured taxonomy, categorizing foundation models based on their modalities and functionalities within the AD domain but also delves into the methods employed in current research.  
    It identifies the gaps between existing foundation models and cutting-edge AD approaches, thereby charting future research directions and proposing a roadmap for bridging these gaps.
}

\maketitle

\newpage
\section{Introduction}\label{sec1}

The integration of deep learning (DL) into autonomous driving (AD) has marked a significant leap in this field, attracting attention from both academic and industrial spheres.
AD systems, equipped with cameras and lidars, mimic human-like decision-making processes.
These systems are fundamentally composed of three key components: perception, prediction, and planning.
Perception, utilizing DL and computer vision algorithms, focuses on object detection and tracking. 
Prediction forecasts the behavior of traffic agents and their interaction with autonomous vehicles. 
Planning, which is typically structured hierarchically, involves making strategic driving decisions, calculating optimal trajectories, and executing vehicle control commands.
%
%
The advent of foundation models, particularly renowned in natural language processing and computer vision, has introduced new dimensions to AD research. 
These models are distinct due to their training on extensive web-scale datasets and their massive parameter sizes. 
Given the vast amounts of data generated by autonomous vehicle services and advancements in AI, including NLP and AI-generated content (AIGC), there is a growing curiosity about the potential of foundation models in AD.
These models could be instrumental in performing a range of AD tasks, such as object detection, scene understanding, and decision-making, with a level of intelligence akin to human drivers.

Foundation models address several challenges in AD. 
Conventionally, AD models are trained in a supervised manner, dependent on manually annotated data that often lack diversity, limiting their adaptability.
Foundation models, however, show superior generalization capabilities due to their training on diverse, web-scale data.  
They can potentially replace the complex heuristic rule-based systems in planning with their reasoning capabilities and knowledge derived from extensive pre-training. 
For example, LLM has reasoning capability and common sense driving knowledge acquired from the pre-training dataset, which can potentially replace heuristic rule-based planning systems, which require complex engineering effort of hand-crafted rules in software codes and debugging on corner cases. 
Generative models within this domain can create realistic traffic scenarios for simulation, essential for testing safety and reliability in rare or challenging situations. 
Moreover, foundation models contribute to making AD technology more user-centric, with language models understanding and executing user commands in natural language.

\begin{figure}
    \centering
    \includegraphics[width=1\linewidth]{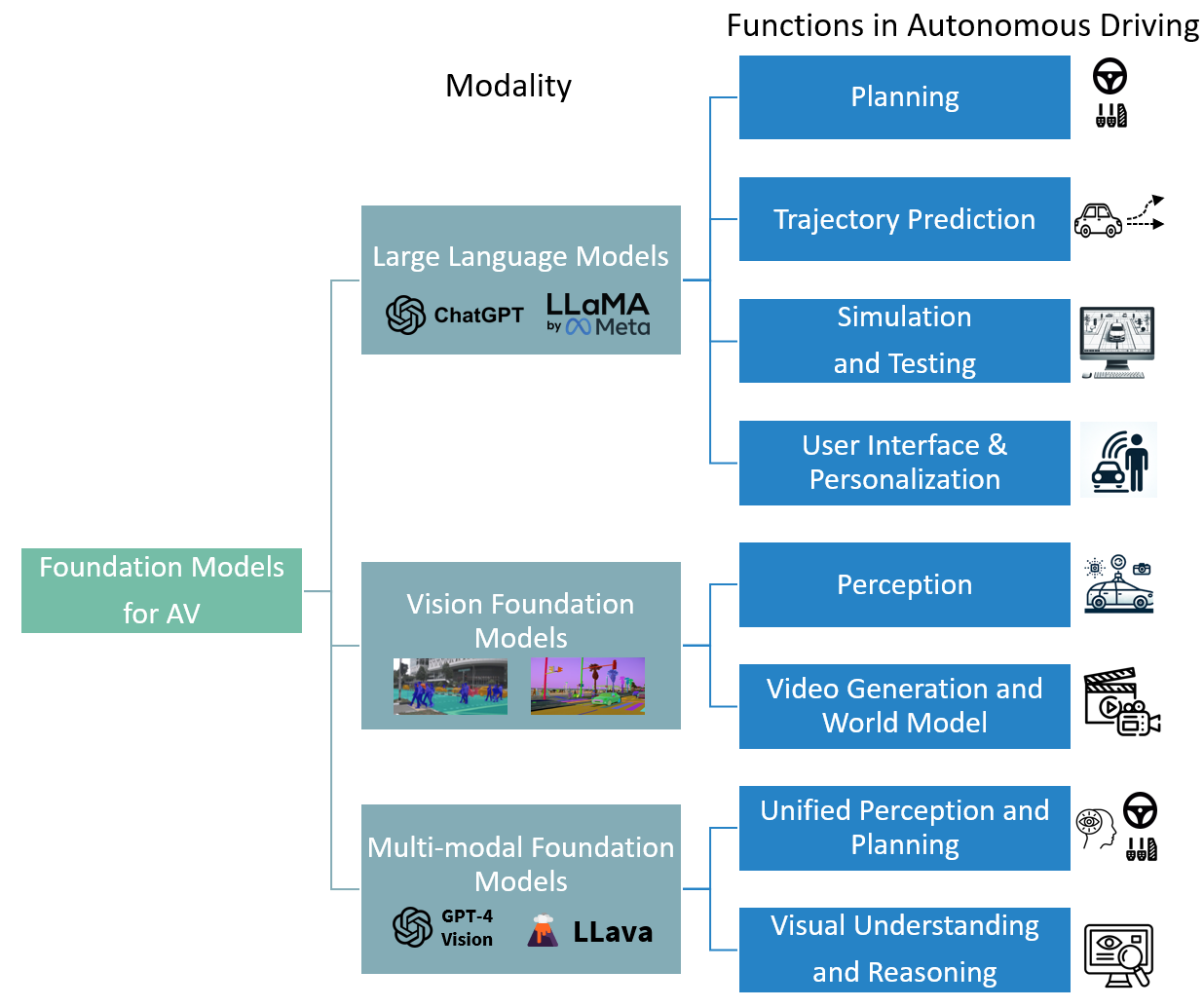}
    \caption{Taxonomy of foundation models for autonomous driving. It delineates the categorization of foundation models according to their modalities, such as Large Language Models, Vision Foundation Models, and Multi-modal Foundation Models, and correlates them with their respective functions in autonomous driving.}
    \label{fig:taxonomy}
\end{figure}

Despite considerable research in applying foundation models to AD, there are notable limitations and gaps in real-world application.
Our survey aims to provide a systematic review and propose future research directions. 
There are two surveys related to foundation models for autonomous driving: LLM4Drive \cite{yang2023llm4drive} is more focused on large language models. \cite{huang2023applications} has a good breadth of summary of applications of foundation models in autonomous driving, mainly in simulation, data annotation, and planning. 
We expand upon existing surveys by covering vision foundation models and multi-modal foundation models, analyzing their applications in prediction and perception tasks. 
%
This comprehensive approach includes detailed examinations of technical aspects, such as pre-trained models and methods, and identifies future research opportunities.
Innovatively, we propose a taxonomy categorizing foundation models in AD based on modalities and functions, as shown in Figure \ref{fig:taxonomy}.
In the following sections, we will explore the application of various foundation models, including large language models, vision foundation models, and multi-modal foundation models, in the context of AD.

\section{Large Language Models in AD}
\subsection{Overview}
LLMs, originally transformative in NLP, are now driving innovations in AD.
Bidirectional Transformers (BERT) \cite{devlin2018bert} pioneered foundation models in NLP, leveraging Transformer architecture for understanding language semantics.
%
%
This pre-trained model can be fine-tuned on specific data-sets, and achieve state-of-the-art results in a wide range of tasks. 
Following this, OpenAI's generative pre-trained transformer (GPT) series \cite{radford2018improving}, including GPT-4, demonstrated remarkable NLP capabilities, attributed to training on extensive datasets. 
%
%
Later GPT models, including ChatGPT, GPT-4 \cite{achiam2023gpt4technical} are trained using billions of parameters and crawled web data with trillions of words, and achieve strong performance on many NLP tasks, including translation, text summarization, question-answering. It also demonstrates one-shot and few-shot reasoning capabilities to learn new skills from the context.
More and more researchers have started to apply these reasoning, understanding, and in-context learning capabilities to address challenges in AD.

\subsection{Applications in AD}
\subsubsection{Reasoning and Planning}
\begin{figure}
    \centering
    \includegraphics[width=\linewidth]{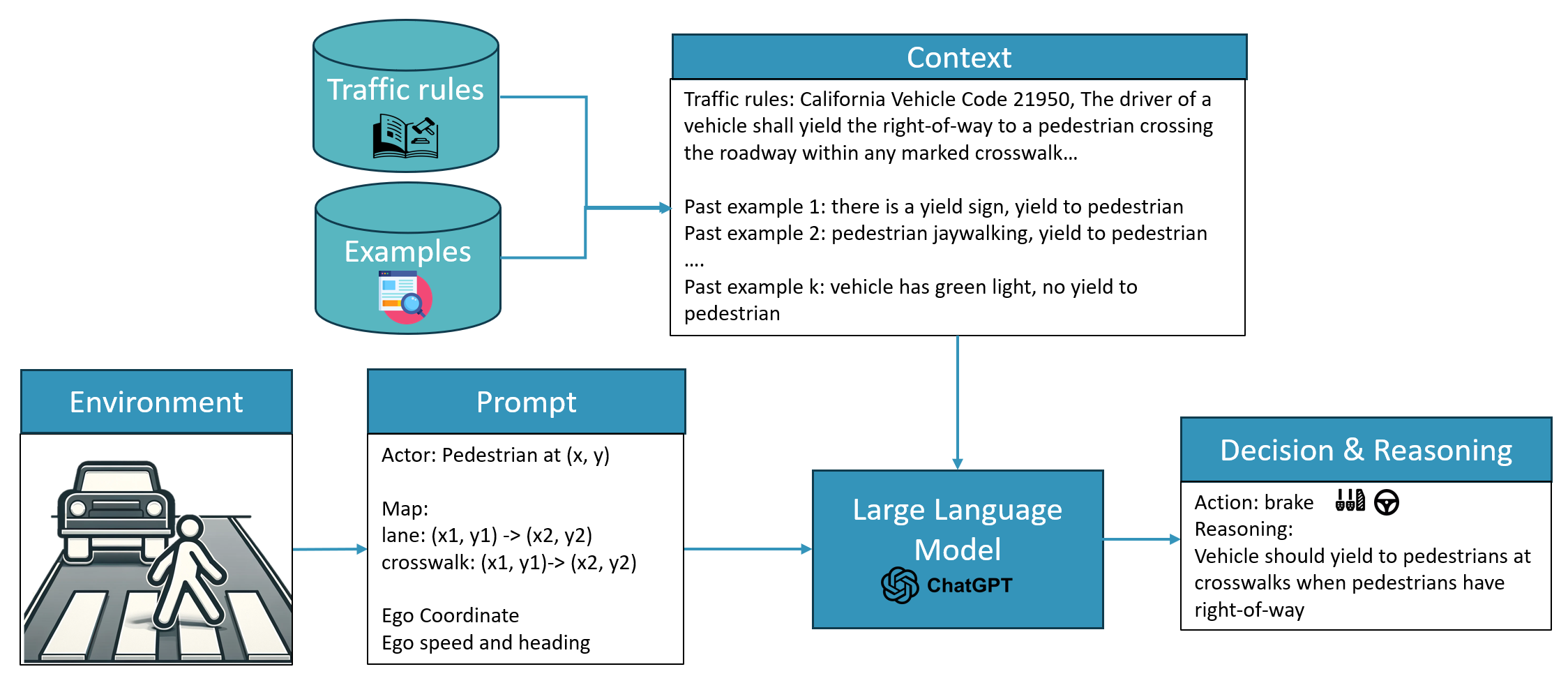}
    \caption{Common pattern of LLM pipelines for autonomous driving, showcasing the integration of textual environment descriptions and LLM reasoning to inform driving decisions.}
    \label{fig:llmpattern}
\end{figure}

The decision-making process in AD closely parallels human reasoning, necessitating the interpretation of environmental cues to make safe and comfortable driving decisions. 
LLMs, through their training on diverse web data, have assimilated common-sense knowledge pertinent to driving, drawing from a plethora of sources including web forums and official government websites. 
This wealth of information enables LLMs to engage in the nuanced decision-making required for AD.
One method for harnessing LLMs in AD involves presenting them with detailed textual descriptions of the driving environment, prompting them to propose driving decisions or control commands. 
This process, as illustrated in Figure \ref{fig:llmpattern}, typically encompasses comprehensive prompts detailing agent states, such as coordinates, speed, and past trajectories, the vehicle's state \textit{i}.\textit{e}, velocity, and acceleration, and map specifics including traffic lights, lane information, and intended route).
For enhanced interaction understanding, LLMs can also be directed to provide reasoning along with their responses. 
For instance, the GPT driver \cite{mao2023gptdriver} not only recommends vehicle actions but also elucidates the rationale behind these suggestions, significantly enhancing the transparency and explainability of autonomous driving decisions. 
This approach, exemplified by Driving with LLMs \cite{chen2023driving}, enhances the explainability of autonomous driving decisions.
Similarly, the ``Receive, Reason, and React'' approach \cite{cui2023receive} instructs LLM agents to assess lane occupancy and evaluate the safety of potential actions, thereby fostering a deeper comprehension of dynamic driving scenarios.
These methods not only leverage LLMs' inherent ability to understand complex scenarios but also employ their reasoning capabilities to simulate human-like decision-making processes. 
Through the integration of detailed environmental descriptions and strategic prompts, LLMs contribute significantly to the planning and reasoning aspects of AD, offering insights and decisions that mirror human judgment and expertise.

\subsubsection{Prediction}
Prediction forecasts traffic participants' future trajectories, intents, and possible interactions with the ego vehicle. The common deep learning-based models are based on rasterized or vector images of the traffic scene, which encode spatial information. However, it is still challenging to accurately predict highly interactive scenes, which requires reasoning and semantic information, for example, right-of-ways, vehicles' turning signals, and pedestrians' gestures. The text representation of the scene can provide more semantic information, and better leverage LLM's reasoning capability and common knowledge in the pre-training dataset. There is still not much research applying LLM to trajectory prediction. \cite{keysan2023canyoutext} did an early exploration of LLM's power to make trajectory predictions. They convert the scene representation into text prompts, and use BERT model to generate the text encoding, which is finally fused with image encoding to decode trajectory prediction. Their evaluation shows significant improvement compared with baselines only using image encoding or text encoding.

\subsubsection{User Interface and Personalization}
Autonomous vehicles should be user-friendly and able to follow the instructions from passengers or remote operators. The current Robotaxi remote assistance interface is only used to execute a limited set of pre-defined commands. However, LLM's understanding and interaction capabilities make it possible to let autonomous driving cars to understand human's free-form instruction to better control the autonomous vehicle and satisfy users' personalized requirements. \cite{cui2023receive} explores the LLM-based planner conditioning on personalized commands, e.g. ``driving aggressively" or ``conservatively", and was able to output actions of various speeds and riskiness. \cite{yang2023humancentric} leverages LLM's reasoning abilities, and provides step-by-step rules to decide on response to user commands. The LLM agent is also able to accept or reject user commands based on pre-defined traffic rules and system requirements.

\subsubsection{Simulation and Testing}
LLM can summarize and extract knowledge from existing text data and generate new content, which can facilitate simulation and testing. The ADEPT system \cite{wang2022adept} uses GPT to extract key information from NHTSA accident reports using QA approach, and was able to generate diverse scene code used for simulation and testing. TARGET \cite{deng2023target} system is able to use GPT to translate traffic rules from the natural language to the domain-specific language, which is used for generating testing scenarios.  LCTGen \cite{tan2023lctgen} uses LLM as a powerful interpreter translating user's text query into structured specifications of map lanes and vehicle locations for traffic simulation scenarios.

\subsection{Methods and Techniques}
Researchers use similar techniques in natural language processing to utilize LLM for autonomous driving tasks, such as prompt engineering, in-context and few-shot learning, and reinforcement learning from human feedback (RLHF)\cite{ouyang2022instructgpt}. 

\subsubsection{Prompt Engineering}

Prompt engineering adopts sophisticated designs of input prompts and questions to guide the Large Language Model to generate our desired answers. 

Some papers add traffic rules as pre-prompt to make LLM agent law-compliant.
Driving with LLMs \cite{chen2023driving} has diving rules covering aspects like traffic light transition and left or right driving side. \cite{mao2023language} proposes a module called common-sense module, which stores the rules and instructions for human driving, for example, avoiding collision, and maintaining safety distances.

LanguageMPC \cite{sha2023languagempc} adopts a top-down decision-making system: given different situations, the vehicle has different possible actions. LLM agent is also instructed to identify important agents in the scenario and output attention, weight, and bias matrices to select from pre-defined actions.

Memory modules are also introduced in some papers, which store past driving scenarios. At inference time, the relevant examples are retrieved and added as the context in the prompt, and LLM agent can better leverage few-shot learning capabilities and reflect on the most relevant scenarios. DILU\cite{wen2023dilu} proposes a memory module, which stores text descriptions of driving scenarios in the vector database, and the system can retrieve top-k scenarios for few-shot learning. \cite{mao2023language} has a two-stage retrieval process: the first stage uses k-nearest-neighbor search to retrieve relevant past examples in the database, and the second stage asks LLM to rank these examples.

More papers built complex systems to manage tasks in the prompt generation, which trigger function calls to other modules or sub-systems to obtain required information for decision-making. \cite{mao2023language} has created libraries and function API calls to interact with perception, prediction, and mapping systems so that the LLM can fully leverage all available information. LanguageMPC \cite{sha2023languagempc} uses LangChain to create tools and interfaces needed by LLM to get relevant vehicles, possible situations, and available actions.

\subsubsection{Fine-tuning v.s. In-context Learning}
Fine-tuning and in-context learning are both applied to adapt pre-trained models to autonomous driving. Fine-tuning re-trains the model parameters on smaller domain-specific datasets, while in-context learning or few-shot learning leverages LLM's knowledge and reasoning ability to learn from given examples in the input prompt. Most papers are focused on in-context learning, but only a few papers utilize fine-tuning. Researchers have mixed results on which one is the better: \cite{mao2023language} compared both approaches and found that few-shot learning is slightly more effective. GPT-Driver \cite{mao2023gptdriver} has a different conclusion that using OpenAI fine-tuning performs significantly better than few-shot learning. \cite{chen2023driving} also compared training from scratch and fine-tuning approaches, and found using the pre-trained LLaMA model with LoRA-based fine-tuning can perform better than training from scratch.

\subsubsection{Reinforcement Learning and Human Feedback}

DILU \cite{wen2023dilu} proposes reflection modules, which store good driving examples and bad driving examples with human corrections to enhance its reasoning capabilities further. In this way, the LLM can learn to reason about what action is safe and unsafe and continuously reflect on a large amount of past driving experiences. Surreal Driver \cite{jin2023surrealdriver} interviewed 24 drivers and used their descriptions of driving behavior as chain-of-thought prompts to develop a ‘coach agent’ module, which can instruct the LLM model to have a human-like driving style. Incorporating Voice Instructions \cite{wang2022incorporating} uses instructions from human coaches, and builds a taxonomy of natural language instructions of action, reward, and reasoning, which are used to train a deep reinforcement learning-based autonomous driving agent.

\subsection{Limitations and Future Directions}

\subsubsection{Hallucination and Harmfulness}
Hallucination is a big challenge in LLM to avoid, and state-of-the-art large language models can still produce misleading and false information, not only in text-based models but also in scenarios like multimodality-based tasks. Hallucination happens when a model generates output that contains fictional, misleading, or entirely fabricated details, facts, or claims, instead of delivering reliable and truthful information. When talking about its relationship with AD, most methods proposed in existing papers still require parsing driving actions from LLM's response. When given an unseen scenario, the LLM model can still produce unhelpful or wrong driving decisions. Autonomous driving is a safety-critical application, which has much higher reliability and safety requirements than chat-bots. According to evaluation result \cite{mao2023gptdriver}, the LLM model for autonomous driving has a 0.44\% collision rate, higher than other methods. \cite{chen2023driving} proposes a method to reduce hallucination by asking questions without enough information to make decisions, and instructs LLM to answer ``I don't know". The pre-trained LLM may also include harmful content, for example, aggressive driving and speeding. More human-in-the-loop training and alignment (like RLHF \cite{kaufmann2023survey} and DPO \cite{rafailov2024direct}) can reduce hallucinations and harmful driving decisions.

\subsubsection{Latency and Efficiency}
Large language models often suffer from high latency, and generating detailed driving decisions can exhaust the latency budget of limited compute resources in the car. It takes several seconds for inference according to \cite{jin2023surrealdriver}. LLMs with billions of parameters can consume over 100GB of memory, which might interfere with other critical modules in autonomous driving vehicles.  
Currently, many methods have been proposed to boost the inference speed for LLM-based decoder models, including model compression and knowledge distillation. A new attention structure like PagedAttention\cite{kwon2023efficient} has been proposed for fast inference in the generative model. Quantization methods like GPTQ\cite{frantar2022gptq}, AWQ\cite{lin2023awq},  SqueezeLLM\cite{kim2023squeezellm} has been designed to achieve around 2.1x to 2.3x speedup compared with the baseline. Distillation methods like SLidR \cite{sautier2022image}, ST-SLiDR\cite{mahmoud2023self} have been proposed for LiDAR style input on detection and segmentation tasks.

\subsubsection{Dependency on Perception System\label{sec:dep_perc}}
Despite the supreme reasoning capability of LLM, the environment description still depends on the upstream perception module. The driving decisions can go wrong and cause critical accidents with minor errors in environmental inputs. For example, \cite{mao2023language} shows failure cases when upstream heading data has errors. LLM also needs to be better adapted to perception models and make better decisions when there are errors and uncertainty.

\subsubsection{Sim to Real Gap}
Most of the research is done in simulated environments, and driving scenarios are much simpler than real-world environments. A lot of engineering and human detailed annotation efforts are needed for prompt engineering to cover all scenarios in the real world, for example, the model knows how to yield to humans, but is probably not good at handling interaction with small animals.

\subsection{Summary}
\begin{figure}
    \centering
    \includegraphics[width=\linewidth]{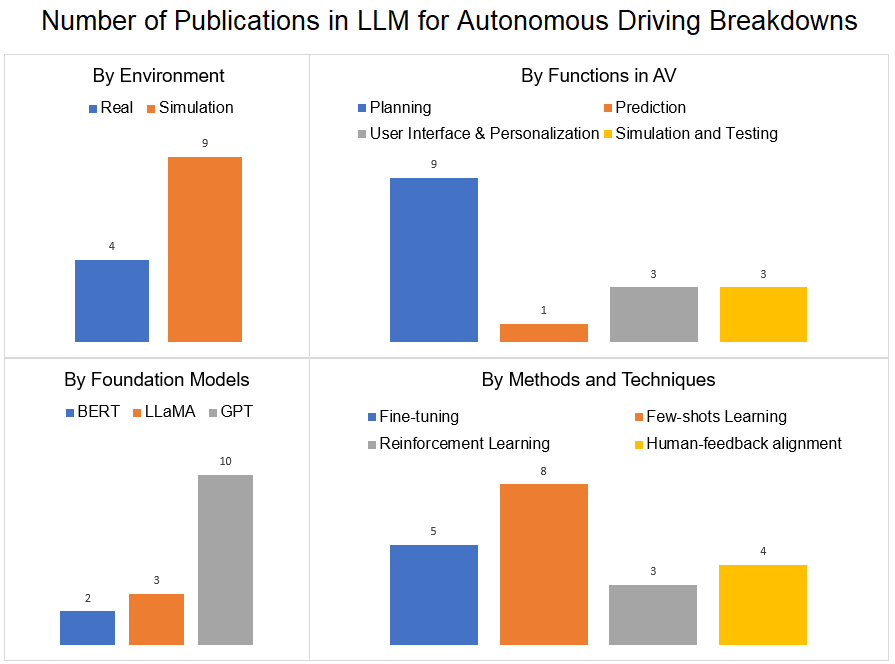}
    \caption{Summary of publications in LLM for autonomous driving.}
    \label{fig:llm-summary}
\end{figure}

The publications in LLM are summarized in Figure \ref{fig:llm-summary}. We propose more fine-grained classifications by environments(real or sim), functions in autonomous driving, foundation models, and techniques used in the research.

\section{Vision Foundation Model}

Vision foundation models have achieved great success in multiple computer vision tasks, such as object detection and segmentation. DINO \cite{caron2021emerging} uses vision-transformer architecture, and is trained in a self-supervised manner, predicting global image features given local image patches. DINOV2 \cite{oquab2023dinov2} scales the training with one billion parameters and a diversely curated dataset of 1.2 billion images and achieves state-of-the-art results in multiple tasks. Segment-anything model \cite{kirillov2023segment} is a foundation model for image-segmentation. The model is trained with different types of prompts (points, boxes, or texts) to generate segmentation masks. Trained with billions of segmentation masks in the dataset, the model shows zero-shot transfer capability to segment new objects given the appropriate prompt.

\begin{figure}
    \centering
    \includegraphics[width=\linewidth]{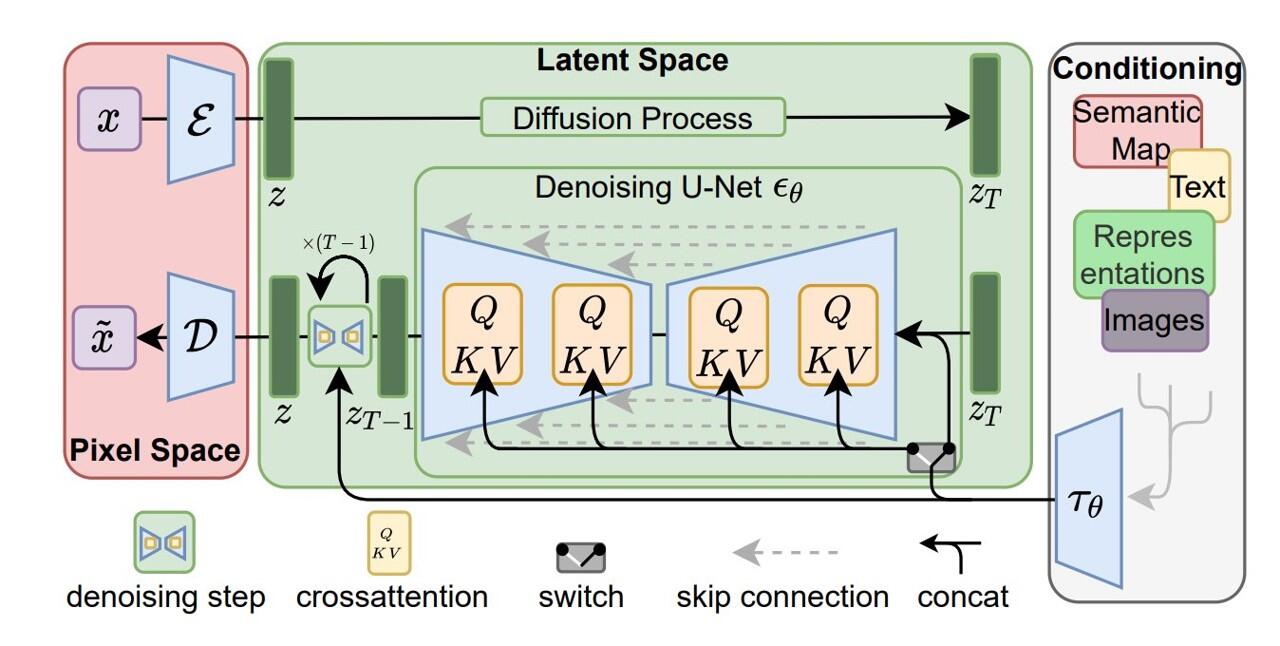}
    \caption{Stable Diffusin}
    \label{fig:stable_diffusion}
\end{figure}

Diffusion model \cite{sohl2015diffusion} is a generative foundation model widely used for image generation. 
In recent years, the diffusion model \cite{yang2023diffusion} has gained significant acclaim in the field of image synthesis. This model generates images by starting with Gaussian noise and progressively refining it through a series of denoising steps. The methodology is grounded in solid physical principles, encompassing both a diffusion phase and a reverse phase. During the diffusion phase, an image is incrementally transformed into a Gaussian distribution by repeatedly adding random Gaussian noise. Conversely, the reverse phase involves reconstructing the original image from this distribution through multiple denoising iterations.
The diffusion model iteratively adds noise to the image and applies a reverse diffusion process to restore the image. To generate the image, we can sample from the learned distribution and restore highly realistic images from random noises. 
Latent diffusion models (LDM) (also known as the stable-diffusion model) \cite{rombach2022high} are a type of diffusion model that focuses on the distribution of the latent space of images and have recently demonstrated outstanding performance in image synthesis. An LDM comprises two components: an autoencoder and a diffusion model, as illustrated in Figure \ref{fig:stable_diffusion}. The autoencoder is responsible for compressing and reconstructing images, utilizing an encoder and a decoder. The encoder transforms the image into a lower-dimensional latent space, while the decoder reconstructs the original image from this latent space. Subsequently, the latent generative model is trained to replicate a fixed forward Markov chain using DDPMs \cite{ho2020denoising}. Stable-Diffusion \cite{rombach2022high} model uses VAE \cite{kingma2013vae} to encode images to latent representation and use UNet \cite{ronneberger2015unet} to decode from latent variable to pixel-wise images. It also has an optional text encoder and applies the cross-attention mechanism to generate images conditional on prompts (text description or other images). 

DALL-E \cite{ramesh2021dalle} model was trained with billions of image and text pairs and uses stable diffusion to generate high-fidelity images and creative arts following human instructions. DALL-E 2 \cite{ramesh2022hierarchical}, an extension of DALL-E \cite{ramesh2021dalle}, integrates a CLIP encoder with a diffusion decoder to handle both image generation and editing tasks, as depicted in Fig. 19. Unlike Imagen, DALL-E 2 utilizes a prior network to translate between text embeddings and image embeddings. Building on this, DALL-E 3 focuses on enhancing prompt adherence and caption quality. It first trains a robust image captioner capable of generating detailed and accurate image descriptions, and then uses this captioner to produce even more refined and detailed captions.
There is growing interest in the application of vision foundation models in autonomous driving, mainly for 3D perception and video generation tasks.

\subsection{Perception}

SAM3D \cite{zhang2023sam3d} applies SAM(Segment-anything model) to 3D object detection in autonomous driving. Lidar point clouds are projected to BEV(bird-eye-view) images, and it uses 32x32 mesh grids to generate point prompts to detect masks for foreground objects. It leverages the SAM model's zero-shot transfer capability to generate segmentation masks and 2D boxes. Then it uses vertical attributes of those lidar points inside 2D boxes to generate 3D boxes. However, the Waymo Open Dataset evaluation shows the average-precision metrics are still far from existing state-of-the-art 3D object detection models. They observed that SAM trained foundation model can not handle those sparse and noisy points very well, and often results in false negatives for distant objects.

SAM is applied to domain adaptation for 3D segmentation tasks, leveraging the SAM model's feature space which contains more semantic information and generalization capability. \cite{peng2023adaptsam} proposes SAM-guided feature alignment, learning unified representation of 3D point cloud features from different domains. It uses the SAM feature extractor to generate the camera image's feature embedding and projects 3D point clouds into camera images to obtain SAM features. The training process optimizes the alignment loss so that 3D features from different domains have a unified representation in SAM's feature space. This approach achieves state-of-the-art performance in 3D segmentation in multiple domain switching datasets, e.g. different cities, weathers, and lidar devices.

SAM and Grounding-DINO\cite{liu2023groundingdino} are used to create a unified segmentation and tracking framework leveraging temporal consistency between video frames\cite{cheng2023samtrack}. Grounding-DINO is an open-set object detector that takes input from text descriptions of objects and outputs the corresponding bounding boxes. Given the text prompts of object classes related to autonomous driving, it can detect objects in video frames and generate bounding boxes of vehicles and pedestrians. SAM model further takes these boxes as prompts and generates segmentation masks for detected objects. The resulting masks of objects are then passed to the downstream tracker, which compares the masks from continuous frames to determine if there are new objects.

\subsection{Video Generation and World Model}
The foundation models, especially generative models and world models can generate realistic virtual driving scenes, which can be used for autonomous driving simulation. Many researchers have started to apply diffusion models to autonomous driving for realistic scene generation. The video generation problem is often formulated as a world model:  given the current world state, conditioning on environment input, the model predicts the next world state and uses diffusion to decode highly realistic driving scenes.

\begin{figure}
    \centering
    \includegraphics[width=1\linewidth]{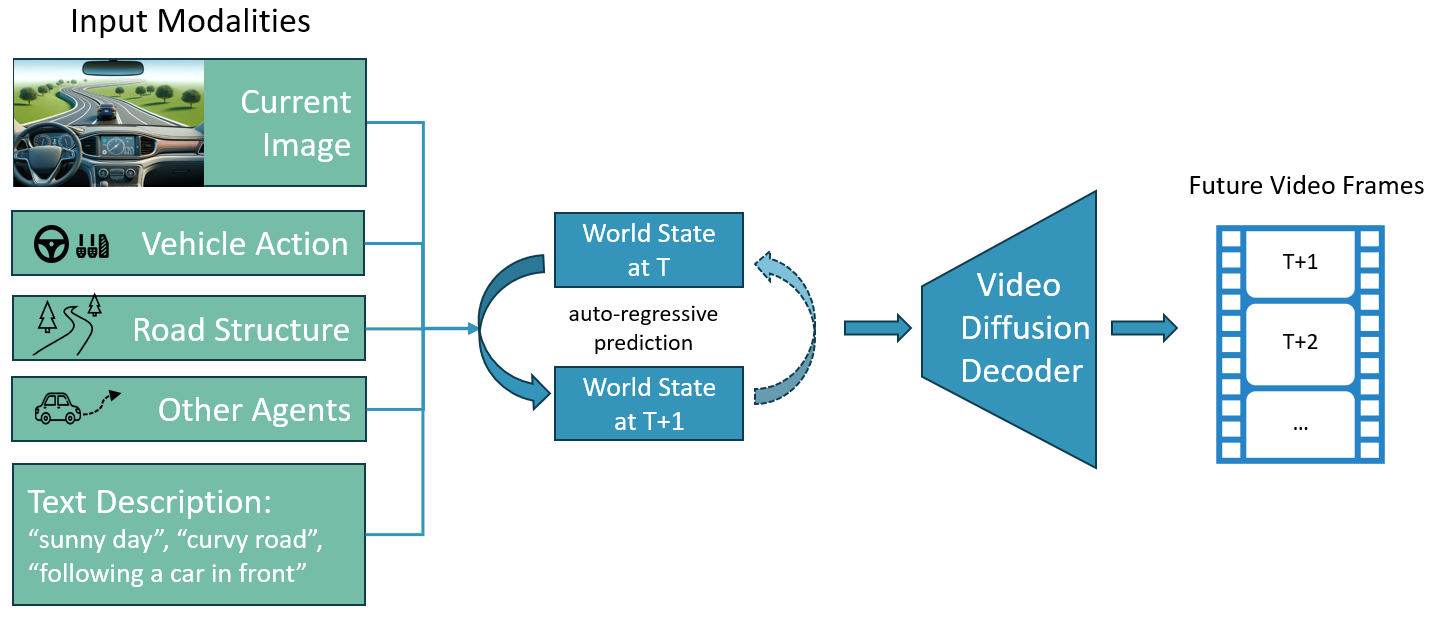}
    \caption{Video generation pipeline for AD.}
    \label{fig:enter-label}
\end{figure}

GAIA-1\cite{hu2023gaia1} is developed by Wayve to generate realistic driving videos. The world model uses camera images, text descriptions, and vehicle control signals as input tokens and predicts the next frame. The paper uses pre-trained DINO\cite{caron2021emerging} model's embedding and cosine similarity loss to distill more semantic knowledge to image token embedding. They use the video diffusion model\cite{ho2022video} to decode high-fidelity driving scenes from the predicted image token. There are two separate tasks to train the diffusion model: image generation and video generation. The image generation task helps the decoder generate high-quality images, while the video generation task uses temporal attention to generate temporally consistent video frames. The generated video follows high-level real-world constraints and has realistic scene dynamics, such as the object's location, interactions, traffic rules, and road structures. The video also shows diversity and creativity, which have realistic possible outcomes conditioned on different text descriptions and the ego vehicle's action.

DriveDreamer 
 \cite{wang2023drivedreamer} also uses the world model and diffusion model to generate video for autonomous driving. In addition to images, text descriptions, and vehicle actions, the model also uses more structural traffic information as input, such as HDMap and object 3D boxes, so that the model can better understand higher-level structural constraints of traffic scenes. The model training has two stages: the first stage is video generation using the diffusion model conditioned on structured traffic information. It was built on a pre-trained Stable-Diffusion model\cite{rombach2022high} with parameters frozen. In the second stage, the model is trained with both future video prediction tasks and action prediction tasks to better learn future prediction and interactions between objects.

\cite{zhang2023discretediffusion} built a point cloud-based world model that achieves SOTA performance in point cloud forecasting tasks. They propose a VQVAE-like \cite{oord2017neural} tokenizer to represent 3D point clouds as latent BEV tokens and use discrete diffusion to forecast future point clouds given the past BEV tokens and ego vehicle's actions tokens.

\subsection{Limitations and Future Directions}
The current state-of-the-art foundation model like SAM doesn't have good enough zero-shot transfer ability for 3D autonomous driving perception tasks, such as object detection, and segmentation. Autonomous driving perception relies on multiple cameras, lidars, and sensor fusions to obtain the highest accuracy object detection result, which is much different from image datasets randomly collected from the web. The scale of current public datasets for autonomous driving perception tasks is still not large enough to train a foundational model and cover all possible long-tail scenarios. Despite the limitation, the existing 2D vision foundation models can serve as useful feature extractors for knowledge distillation, which helps models better incorporate semantic information. In the domain of video generation and forecasting tasks, we have already seen promising progress leveraging existing diffusion models for video generation and point cloud forecasting, which can be further applied to creating high-fidelity scenarios for autonomous driving simulation and testing.

\section{Multi-modal Foundation Models}

Multi-modal foundation models benefit more by taking input data from multiple modalities, e.g. sounds, images, and video, to perform more complex tasks, e.g. generating text from images, analyzing and reasoning with visual inputs.

One of the most well-known multi-modal foundation models is CLIP\cite{radford2021clip}. The model is pre-trained using the contrastive pre-training method. The inputs are noisy images and text pairs, and the model is trained to predict if the given image and text are a correct pair. The model is trained to maximize the cosine similarity of embedding from the image encoder and text encoder. The CLIP model shows zero-shot transfer ability for other computer vision tasks, such as image classification, and predicting the correct text description of the class without supervised training.

Multi-modal foundation models, like LLaVA\cite{liu2023visualllava}, LISA\cite{lai2023lisa}, and CogVLM\cite{wang2023cogvlm} can be used for the general-purpose visual AI agent, which demonstrates superior performance in vision tasks, such as object segmentation, detection, localization, and spatial reasoning. Video-LLaMA\cite{zhang2023videollama} can further perceive video and audio data, which may help autonomous vehicles better understand the world from temporal images and audio sequences.

Multi-modal foundation model is also used for robot learning, which leverages the robot's action as a new modality to create more general-purpose agents that can perform real-world tasks. DeepMind proposed a vision-language-action model\cite{brohan2023rt} trained on text and images from the web and learned to output control commands to complete real-world object manipulation tasks.

Transferring general knowledge from large-scale pre-training datasets to autonomous driving, the multi-modal foundation models can be used for object detection, visual understanding, and spatial reasoning, which enables more powerful applications in autonomous driving.

\subsection{Visual Understanding and Reasoning}
Traditional object detection or classification models are not enough for autonomous driving, because we need better semantic understanding and visual reasoning of the scene, for example, identifying risky objects, and understanding the intents of traffic participants. Most of the existing deep learning-based prediction and planning models are dark-box models, which have poor explainability and debuggability when accidents or discomfort events happen. With the help of the multi-modal foundation models, we can generate explanations and the reasoning process of the model to better investigate the issues.

To further improve the perception system, HiLM-D\cite{ding2023hilmd} utilizes multi-modal foundation models for ROLISP(Risk Object Localization and Intention and Suggestion Prediction). It uses natural language to identify risky objects from camera images and provide suggestions on the ego vehicle's actions. To overcome the drawback of missing small objects, it proposes a pipeline with both high-resolution and low-resolution branches. The low-resolution reasoning branch is used to understand high-level information and identify risk objects from continuous video frames; The high-resolution perception branch enables further refinement of object detection and localization quality. Their model backbone uses the pre-trained visual encoder and LLM weights following BLIP2\cite{li2023blip2}.

Talk2BEV\cite{dewangan2023talk2bev} proposes an innovative bird’s-eye view (BEV) representation of the scene fusing both visual and semantic information. The pipeline first generates the BEV map from image and lidar data and uses general-purpose visual-language foundation models to add more detailed text descriptions of cropped images of objects. The JSON text representation of the BEV map is then passed to general-purpose LLM to perform Visual QA, which covers spatial and visual reasoning tasks. The result shows a good understanding of detailed instance attributes and also higher-level intent of objects, and the ability to provide free-formed advice on the ego vehicle's actions.

LiDAR-LLM\cite{yang2023lidarllm} uses a novel approach that combines point cloud data with the advanced reasoning abilities of Large Language Models to interpret real-world 3D environments and achieves excellent performance in 3D captioning, grounding, and QA tasks. The model employs a unique three-stage training and a View-Aware Transformer(VAT) to align 3D data with text embedding, enhancing spatial comprehension. Their examples show the model can understand the traffic scenes and provide suggestions for autonomous driving planning tasks. 

\cite{atakishiyev2023explaining} focus on the the explainability of vehicle's actions using a visual QA approach. They collected driving videos in simulated environments from 5 different action categories(like going straight and turning left) and used manually labeled explanations of actions to train the model. The model was able to explain the driving decision based on road geometry and clearance of obstacles. They find it promising to apply state-of-the-art multi-modal foundation models to generate structured explanations of vehicle actions.

\subsection{Unified Perception and Planning}

\cite{wen2023gpt4road} performed an early exploration of GPT-4Vision\cite{achiam2023gpt4technical}'s application in perception and planning tasks, and evaluated its capabilities in several scenarios. It shows that GPT-4Vision can understand weather, traffic signs, and traffic lights and identify traffic participants in the scene. It can also provide more detailed semantic descriptions of these objects, such as vehicle rear lights, intents like U-turn, and detailed vehicle types(e.g. cement mixer truck, trailer, and SUV). It also shows the foundation model's potential for understanding point cloud data, GPT-4V can identify vehicles from point cloud contours projected in BEV images. They also evaluated the model's performance on planning tasks. Given the traffic scenario, GPT4-V is asked to describe its observation and decision on the vehicle's action. The results show good interaction with other traffic participants and compliance with the traffic rules and common sense, e.g. following the car at a safety distance, yielding to cyclists at a crosswalk, remaining stopped until the light turns green. It can even handle some long-tail scenarios very well, such as the gated parking lot.

Instruction tuning is used to better adapt general-purpose multi-modal foundation models to autonomous driving tasks. DriveGPT4 \cite{xu2023drivegpt4} created an instruction-following dataset, where ChatGPT, YOLOV8 \cite{reis2023yolov8} and ground truth vehicle control signals from BBD-X dataset \cite{kim2018textual} are used to generate question and answers about common objects detections, spatial relations, traffic light signals, the ego vehicle's actions.  Following LLaVA, it used the pre-trained CLIP\cite{radford2021clip} encoder and LLM weights and fine-tuned the model with their instruction-following dataset specifically designed for autonomous driving. They were able to build an end-to-end interpretable autonomous driving system, which is able to have a good understanding of the surrounding environment and make decisions on vehicle actions with jurisdictions and lower-level control commands. 

\subsection{Limitations and Future Directions}
The multi-modal foundation models show capability for spatial and visual reasoning, which is required by autonomous driving tasks. Compared to traditional object detection, classification model trained on the closed-set dataset, the visual reasoning capability and free-formed text description can provide more abundant semantic information, which can solve many long-tail detection problems, such as classification of special vehicles, and understanding of hand signals from the police officers and traffic controllers. The multi-modal foundation models have good generalization capability and can handle some challenging long-tail scenarios very well using common sense, like stopping at a gate with controlled access. Further leveraging its reasoning capability for planning tasks, the vision-language models can be used for unified perception planning and end-to-end autonomous driving.

There are still limitations of multi-foundation models in autonomous driving. \cite{wen2023gpt4road} shows the GPT-4V model still suffers from hallucination and generates unclear responses or false answers in several examples. The model also shows incompetence in utilizing multi-view cameras and lidar data for accurate 3D object detections and localization, because the pre-training dataset only contains 2D images from the web. More domain-specific fine-tuning or pre-training is required to train multi-modal foundation models to better understand point cloud data and sensor fusion to achieve comparable performance of the state-of-the-art perception system.
 
\section{Conclusion and Future Directions}
We have summarized and categorized recent papers applying foundation models to autonomous driving. We build a new taxonomy based on modality and functions in autonomous driving. We have detailed discussions on methods and techniques for adapting foundation models to autonomous driving, e.g. in-context learning, fine-tuning, reinforcement learning, and visual instruction tuning. We also analyze the limitations of foundation models in autonomous driving, e.g. hallucination, latency, and efficiency as well as the domain gap in the dataset, and thereby propose the following research directions:

\begin{itemize}
    \item Domain-specific pre-training or fine-tuning on autonomous driving dataset
    \item Reinforcement Learning, and Human-in-the-loop alignment to improve safety and reduce hallucinations 
    \item Adaptation of 2D foundation models to 3D, e.g. language guided sensor fusion, fine-tuning, or few-shot learning on the 3D dataset 
    \item Latency and memory optimization, model compression, and knowledge distillation for deployment of foundation models to vehicles    
\end{itemize}

We also notice that the dataset is one of the biggest obstacles in the future development of foundation models in autonomous driving. The existing open-sourced dataset\cite{li2024opensourced} for autonomous driving at the scale of 1000 hours, is far less than pre-training datasets used for state-of-the-art LLMs. The web dataset used for existing foundation models doesn't leverage all modalities required by autonomous driving, such as lidar and surround cameras. The web data domain is also quite different from the real driving scenes.

\begin{figure}
    \centering
    \includegraphics[width=1\linewidth]{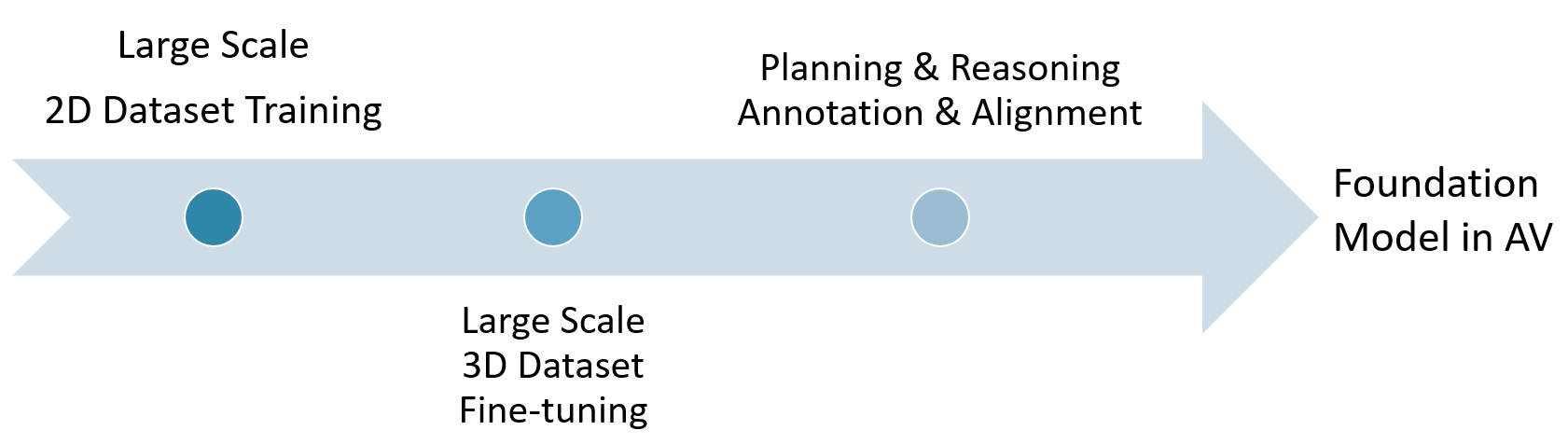}
    \caption{Road map of foundation models in AD.}
    \label{fig:roadmap}
\end{figure}

We propose the longer-term future road map in Figure \ref{fig:roadmap}. In the first stage, we can collect a large-scale 2D dataset that can cover all data distribution, diversity, and complexity of driving scenes in the real-world environment for pre-training or fine-tuning. Most vehicles can be equipped with front cameras to collect the data in different cities, at various times of the day. In the second stage, we can use smaller but higher-quality 3D datasets with lidar to improve the foundation model's 3D perception and reasoning, for example, we can use existing state-of-the-art 3D object detection models as teachers to fine-tune the foundation model. Finally, we can leverage human driving examples or annotations in planning and reasoning for alignment, reaching the utmost safety goal of autonomous driving.

\subsection*{Conflict of Interest}
The authors declare that they have no known competing financial interests or personal relationships that could have appeared to influence the work reported in this paper.

\subsection*{Availability of Data}
The data that support the findings of this study are available from the corresponding author upon request.

\subsection*{Ethical Considerations}
This study is a literature review and did not involve human participants or animal subjects. Therefore, formal ethics approval was not required.

\subsection*{Funding Information}
This research received no specific grant from any funding agency in the public, commercial, or not-for-profit sectors.

\subsection*{Author Contributions}
Y.Shen conceptualized the study. H. Gao, Z.Wang, Y. Li, and K. Long performed the literature review and analysis. H. Gao and Y. Shen contributed to the interpretation of the results. H. Gao and Y. Shen drafted the manuscript. M. Yang and Y. Shen contributed to the revision of the manuscript and approved the final version.

\bibliography{sn-bibliography}

\end{document}